\documentclass{llncs}
\usepackage{graphicx}
\usepackage[vietnam,english]{babel}
\usepackage[utf8x]{inputenc}
\usepackage[T1]{fontenc}
 
\usepackage{subfigure}
\usepackage{amsmath}
\usepackage{xcolor}
\usepackage{tikz}
\usetikzlibrary{shapes.geometric, arrows}

\tikzstyle{process} = [rectangle, minimum width=3em, minimum height=2em, text centered, draw=black, fill=white!10]
\tikzstyle{decision} = [diamond, text centered, aspect=3.5, draw=black, fill=white!10]
\tikzstyle{startend} = [ellipse, minimum width=2em, minimum height=1em, text centered, draw=black, fill=white!10]

\tikzstyle{arrow} = [thin,->,>=stealth]
\begin{document}

\title{Error Analysis for Vietnamese Named Entity Recognition on Deep Neural Network Models}
\author{Binh An Nguyen \and Kiet Van Nguyen \and
Ngan Luu-Thuy Nguyen}
\authorrunning{Binh An Nguyen et al.} 
\institute{University of Information Technology, \newline Vietnam National University - Ho Chi Minh City, Vietnam\\
\email{nguyenanbinh96@gmail.com, kietnv@uit.edu.vn, ngannlt@uit.edu.vn}}

\maketitle              

\begin{abstract}In recent years, Vietnamese Named Entity Recognition (NER) systems have had a great breakthrough when using Deep Neural Network methods. This paper describes the primary errors of the state-of-the-art NER systems on Vietnamese language. After conducting experiments on BLSTM-CNN-CRF and BLSTM-CRF models with different word embeddings on the Vietnamese NER dataset. This dataset is provided by VLSP in 2016 and used to evaluate most of the current Vietnamese NER systems. We noticed that BLSTM-CNN-CRF gives better results, therefore, we analyze the errors on this model in detail. Our error-analysis results provide us thorough insights in order to increase the performance of NER for the Vietnamese language and improve the quality of the corpus in the future works.
\end{abstract}
\section{Introduction}
Named Entity Recognition (NER) is one of information extraction subtasks that is responsible for detecting entity elements from raw text and can determine the category in which the element belongs, these categories include the names of persons, organizations, locations, expressions of times, quantities, monetary values and percentages.

The problem of NER is described as follow:

\textbf{Input}: A sentence S consists a sequence of $n$ words: $S= w_1,w_2,w_3,…,w_n$ ($w_i$: the $i^{th}$ word)

\textbf{Output}: The sequence of $n$ labels $y_1,y_2,y_3,…,y_n$. Each $y_i$ label represents the category which $w_i$ belongs to.

For example, given a sentence:

\textbf{Input}: 
\foreignlanguage{vietnam}{Giám đốc điều hành Tim Cook của Apple vừa giới thiệu 2 điện thoại iPhone, đồng hồ thông minh mới, lớn hơn ở sự kiện Flint Center, Cupertino.}

(Apple CEO Tim Cook introduces 2 new, larger iPhones, Smart Watch at Cupertino Flint Center event)\footnote{http://sanfrancisco.cbslocal.com/2014/09/09/apple-ceo-tim-cook-introduces-2-new-iphones-at-cupertino-flint-center-event/}

The algorithm will output:

\textbf{Output}: 
\foreignlanguage{vietnam}{⟨O⟩Giám đốc điều hành⟨O⟩ ⟨PER⟩Tim Cook⟨PER⟩ ⟨O⟩của⟨O⟩ ⟨ORG⟩Apple⟨ORG⟩ ⟨O⟩vừa giới thiệu 2 điện thoại iPhone, đồng hồ thông minh mới, lớn hơn ở sự kiện⟨O⟩ ⟨ORG⟩Flint Center⟨ORG⟩, ⟨LOC⟩Cupertino⟨LOC⟩.}

With LOC, PER, ORG is Name of location, person, organization respectively. Note that O means Other (Not a Name entity). We will not denote the O label in the following examples in this article because we only care about name of entities.

In this paper, we analyze common errors of the previous state-of-the-art techniques using Deep Neural Network (DNN) on VLSP Corpus. This may contribute to the later researchers the common errors from the results of these state-of-the-art models, then they can rely on to improve the model.

Section 2 discusses the related works to this paper. We will present a method for evaluating and analyzing the types of errors in Section 3. The data used for testing and analysis of errors will be introduced in Section 4, we also talk about deep neural network methods and pre-trained word embeddings for experimentation in this section. Section 5 will detail the errors and evaluations. In the end is our contribution to improve the above errors.

\section{Related work}

Previously publicly available NER systems do not use DNN, for example, the MITRE Identification Scrubber Toolkit (MIST) \cite{11}, Stanford NER \cite{12}, BANNER \cite{13} and NERsuite \cite{14}. NER systems for Vietnamese language processing used traditional machine learning methods such as Maximum Entropy Markov Model (MEMM), Support Vector Machine (SVM) and Conditional Random Field (CRF). In particular, most of the toolkits for NER task attempted to use MEMM \cite{6}, and CRF \cite{5} to solve this problem.

Nowadays, because of the increase in data, DNN methods are used a lot. They have archived great results when it comes to NER tasks, for example, Guillaume Lample et al with BLSTM-CRF in \cite{4} report 90.94 F1 score, Chiu et al with BLSTM-CNN in \cite{1} got 91.62 F1 score, Xeuzhe Ma and Eduard Hovy with BLSTM-CNN-CRF in \cite{8} achieved F1 score of 91.21, Thai-Hoang Pham and Phuong Le-Hong with BLSTM-CNN-CRF in \cite{16} got 88.59\% F1 score. These DNN models are also the state-of-the-art models.

\section{Error-analysis method}

The results of our analysis experiments are reported in precision and recall over all labels (name of person, location, organization and miscellaneous). The process of analyzing errors has 2 steps:

\begin{itemize}
\item \textbf{Step 1}: We use two state-of-the-art models including BLSTM-CNN-CRF and BLSTM-CRF to train and test on VLSP’s NER corpus. In our experiments, we implement word embeddings as features to the two systems.

\item \textbf{Step 2}: Based on the best results (BLSTM-CNN-CRF), error analysis is performed based on five types of errors (No extraction, No annotation, Wrong range, Wrong tag, Wrong range and tag), in a way similar to \cite{15}, but we analyze on both gold labels and predicted labels (more detail in figure 1 and 2).
\end{itemize}
A token (an entity name maybe contain more than one word) will be extracted as a correct entity by the model if both of the followings are correct:
\begin{enumerate}
\item The length of it (range) is correct: The word beginning and the end is the same as gold data (annotator).
\item The label (tag) of it is correct: The label is the same as in gold data.
\end{enumerate}

If it is not meet two above requirements, it will be the wrong entity (an error). Therefore, we divide the errors into five different types which are described in detail as follows:
\begin{enumerate}
\item \textbf{No extraction}: The error where the model did not extract tokens as a name entity (NE) though the tokens were annotated as a NE.\newline\newline
\textbf{LSTM-CNN-CRF}: {\foreignlanguage{vietnam}{ Việt\_Nam}}\newline
\textbf{Annotator}: {\foreignlanguage{vietnam}{{⟨LOC⟩ Việt\_Nam ⟨LOC⟩}}}\newline

\item \textbf{No annotation}: The error where the model extracted tokens as an NE though the tokens were not annotated as a NE.\newline\newline
\textbf{LSTM-CNN-CRF}:	\foreignlanguage{vietnam}{⟨PER⟩ Châu Âu ⟨PER⟩}\newline
\textbf{Annotator}: 	\foreignlanguage{vietnam}{Châu Âu}\newline

\item \textbf{Wrong range}: The error where the model extracted tokens as an NE and only the range was wrong. (The extracted tokens were partially annotated or they were the part of the annotated tokens).\newline\newline
\textbf{LSTM-CNN-CRF}: 	\foreignlanguage{vietnam}{⟨PER⟩ Ca\_sĩ Nguyễn Văn A ⟨PER⟩}\newline
\textbf{Annotator}: \newline
			\centerline{\foreignlanguage{vietnam}{Ca\_sĩ ⟨PER⟩ Nguyễn Văn A ⟨PER⟩}}\newline

\item \textbf{Wrong tag}: The error where the model extracted tokens as an NE and only the tag type was wrong.\newline\newline
\textbf{LSTM-CNN-CRF}: 	\foreignlanguage{vietnam}{Khám phá ⟨PER⟩ Yangsuri ⟨PER⟩}\newline
\textbf{Annotator}:\newline		\centerline{\foreignlanguage{vietnam}{Khám phá ⟨LOC⟩ Yangsuri ⟨LOC⟩}}\newline

\item \textbf{Wrong range and tag}: The error where the model extracted tokens as an NE and both the range and the tag type were wrong.\newline\newline
\textbf{LSTM-CNN-CRF}: \foreignlanguage{vietnam}{⟨LOC⟩ gian\_hàng Apple ⟨LOC⟩}\newline
\textbf{Annotator}: \newline		\centerline{\foreignlanguage{vietnam}{gian\_hàng ⟨ORG⟩ Apple ⟨ORG⟩}}\newline
\end{enumerate}

We compare the predicted NEs to the gold NEs ($Fig. 1$), if they have the same range, the predicted NE is a correct or \textbf{Wrong tag}. If it has different range with the gold NE, we will see what type of wrong it is. If it does not have any overlap, it is a \textbf{No extraction}. If it has an overlap and the tag is the same at gold NE, it is a \textbf{Wrong range}. Finally, it is a \textbf{Wrong range and tag} if it has an overlap but the tag is different.
The steps in Fig. 2 is the same at Fig. 1 and the different only is we compare the gold NE to the predicted NE, and \textbf{No extraction} type will be \textbf{No annotation}.

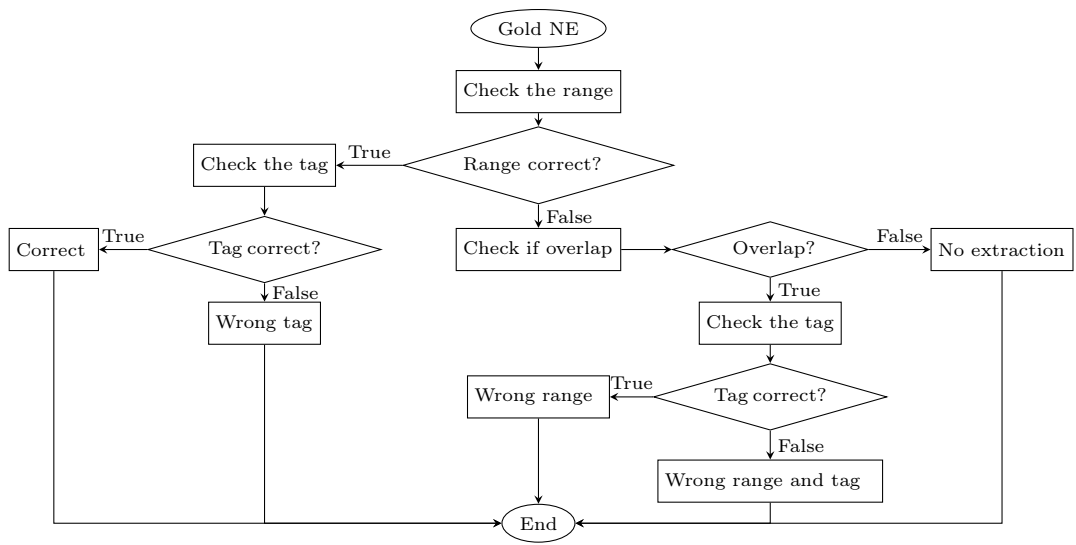
\begin{figure}
\centering
\hfill\\
{\scriptsize \begin{tikzpicture}[node distance=2em]

\node (S) [startend] {Gold NE};

\node (S1) [process, below of=S, yshift= -1em]{\parbox[t][][t]{2cm}{Check the range}};

\node (S2) [decision, below of=S1, yshift=-1.5em]{\parbox[t][][t]{2cm}{Range correct?}};

\node (S3) [process, left of=S2, xshift=-11em, yshift=0em]{\parbox[t][][t]{1.7cm}{Check the tag}};

\node (S4) [decision, below of=S3, yshift=-2em]{\parbox[t][][t]{1.5cm}{Tag correct?}};

\node (S5) [process, left of=S4, xshift=-8em, yshift=0em]{\parbox[t][][t]{1cm}{Correct}};

\node (S6) [process, below of=S4, yshift=-1.5em]{\parbox[t][][t]{1.3cm}{Wrong tag}};

\node (S9) [process, below of=S2, yshift=-2em]{\parbox[t][][t]{2cm}{Check if overlap}};

\node (S7) [decision, right of=S9, xshift=9em]{\parbox[t][][t]{1cm}{Overlap?}};

\node (S8) [process, right of=S7, xshift=9em]{\parbox[t][][t]{1.7cm}{No extraction}};

\node (S10) [process, below of=S7, yshift=-1.5em]{\parbox[t][][t]{1.7cm}{Check the tag}};

\node (S11) [decision, below of=S10, yshift=-1.5em]{\parbox[t][][t]{1.5cm}{Tag correct?}};

\node (S12) [process, below of=S11, yshift=-2em]{\parbox[t][][t]{2.8cm}{Wrong range and tag}};

\node (S13) [process, left of=S11, xshift=-9em]{\parbox[t][][t]{1.7cm}{Wrong range}};

\node (S14) [startend, below of=S13, yshift=-4em] {End};

\draw [arrow] (S) -- (S1);
\draw [arrow] (S1) -- (S2);
\draw [arrow] (S2) -- node[anchor=south]{True}(S3);
\draw [arrow] (S3) -- (S4);
\draw [arrow] (S4) -- node[anchor=south]{True}(S5);
\draw [arrow] (S4) -- node[anchor=west]{False}(S6);
\draw [arrow] (S9) -- (S7);
\draw [arrow] (S2) -- node[anchor=west]{False}(S9);
\draw [arrow] (S7) -- node[anchor=south]{False}(S8);
\draw [arrow] (S7) -- node[anchor=west]{True}(S10);
\draw [arrow] (S10) -- (S11);
\draw [arrow] (S11) -- node[anchor=west]{False}(S12);
\draw [arrow] (S11) -- node[anchor=south]{True}(S13);
\draw [arrow] (S13) -- (S14);
\draw [arrow] (S12) |- (S14);
\draw [arrow] (S8) |- (S14);
\draw [arrow] (S5) |- (S14);
\draw [arrow] (S6) |- (S14);

\end{tikzpicture}}
\caption{Chart flow to analyze errors based on gold labels}
\end{figure}

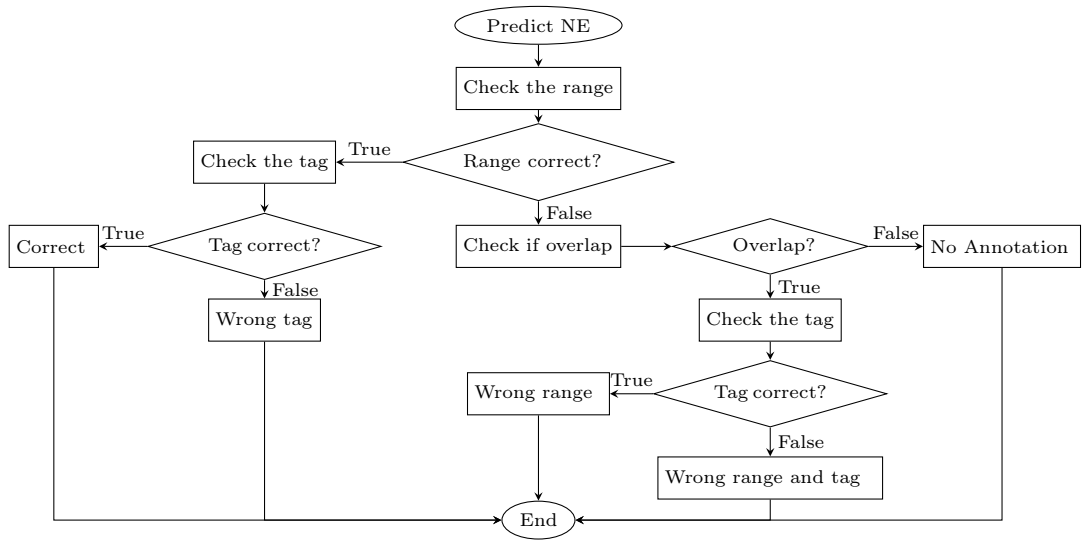
\begin{figure}
\centering
\hfill\\
{\scriptsize \begin{tikzpicture}[node distance=2em]

\node (S) [startend] {Predict NE};

\node (S1) [process, below of=S, yshift= -1em]{\parbox[t][][t]{2cm}{Check the range}};

\node (S2) [decision, below of=S1, yshift=-1.5em]{\parbox[t][][t]{2cm}{Range correct?}};

\node (S3) [process, left of=S2, xshift=-11em, yshift=0em]{\parbox[t][][t]{1.7cm}{Check the tag}};

\node (S4) [decision, below of=S3, yshift=-2em]{\parbox[t][][t]{1.5cm}{Tag correct?}};

\node (S5) [process, left of=S4, xshift=-8em, yshift=0em]{\parbox[t][][t]{1cm}{Correct}};

\node (S6) [process, below of=S4, yshift=-1.5em]{\parbox[t][][t]{1.3cm}{Wrong tag}};

\node (S9) [process, below of=S2, yshift=-2em]{\parbox[t][][t]{2cm}{Check if overlap}};

\node (S7) [decision, right of=S9, xshift=9em]{\parbox[t][][t]{1cm}{Overlap?}};

\node (S8) [process, right of=S7, xshift=9em]{\parbox[t][][t]{1.9cm}{No Annotation}};

\node (S10) [process, below of=S7, yshift=-1.5em]{\parbox[t][][t]{1.7cm}{Check the tag}};

\node (S11) [decision, below of=S10, yshift=-1.5em]{\parbox[t][][t]{1.5cm}{Tag correct?}};

\node (S12) [process, below of=S11, yshift=-2em]{\parbox[t][][t]{2.8cm}{Wrong range and tag}};

\node (S13) [process, left of=S11, xshift=-9em]{\parbox[t][][t]{1.7cm}{Wrong range}};

\node (S14) [startend, below of=S13, yshift=-4em] {End};

\draw [arrow] (S) -- (S1);
\draw [arrow] (S1) -- (S2);
\draw [arrow] (S2) -- node[anchor=south]{True}(S3);
\draw [arrow] (S3) -- (S4);
\draw [arrow] (S4) -- node[anchor=south]{True}(S5);
\draw [arrow] (S4) -- node[anchor=west]{False}(S6);
\draw [arrow] (S9) -- (S7);
\draw [arrow] (S2) -- node[anchor=west]{False}(S9);
\draw [arrow] (S7) -- node[anchor=south]{False}(S8);
\draw [arrow] (S7) -- node[anchor=west]{True}(S10);
\draw [arrow] (S10) -- (S11);
\draw [arrow] (S11) -- node[anchor=west]{False}(S12);
\draw [arrow] (S11) -- node[anchor=south]{True}(S13);
\draw [arrow] (S13) -- (S14);
\draw [arrow] (S12) |- (S14);
\draw [arrow] (S8) |- (S14);
\draw [arrow] (S5) |- (S14);
\draw [arrow] (S6) |- (S14);

\end{tikzpicture}}
\caption{Chart flow to analyze errors based on predicted labels}
\end{figure}

\section{Data and model}
\subsection{Data sets}

To conduct error analysis of the model, we used the corpus which are provided by VLSP 2016 - Named Entity Recognition\footnote{More detail in http://vlsp.org.vn/vlsp2016/eval/ner}. The dataset contains four different types of label: Location (LOC), Person (PER), Organization (ORG) and Miscellaneous - Name of an entity that do not belong to 3 types above (Table \ref{tab:1}). Although the corpus has more information about the POS and chunks, but we do not use them as features in our model.

\begin{table}
\begin{center}
\caption{Number type of each tags in the corpus}
\begin{tabular}{llr}
\hline\rule{0pt}{12pt}Tags & Number of tag & \% \\[2pt]
\hline\rule{0pt}{12pt}Person & 1294 & 43.22 \\
Location	&1379&	46.06\\
Organization&	274&	9.15\\
MISC	&49	&1.64\\
All Tags&	2994	&100\\[2pt]
\end{tabular}
\label{tab:1}
\end{center}
\end{table}

There are two folders with 267 text files of training data and 45 text files of test data. They all have their own format. We take 21 first text files and 22 last text files and 22 sentences of the $22^{th}$ text file and 55 sentences of the $245^{th}$ text file to be a development data. The remaining files are going to be the training data. The test file is the same at the file VSLP gave. Finally, we have 3 text files only based on the CoNLL 2003 format: train, dev and test.
\subsection{Pre-trained word Embeddings}

We use the word embeddings for Vietnamese that created by Kyubyong Park\footnote{The pre-trained word vector of 30+ languages are available at https://github.com/Kyubyong/wordvectors}  and Edouard Grave at al\footnote{The pre-trained word vector of 294 languages are available at https://github.com/facebookresearch/fastText/blob/master/pretrained-vectors.mdh}:\newline
\textbf{Kyubyong Park}: In his project, he uses two methods including fastText\footnote{https://research.fb.com/fasttext/}  and word2vec\footnote{https://code.google.com/archive/p/word2vec/}  to generate word embeddings from wikipedia database backup dumps\footnote{wikipedia database backup dumps: https://dumps.wikimedia.org/backup-index.html}. His word embedding is the vector of 100 dimension and it has about 10k words.\newline
\textbf{Edouard Grave et al} \cite{17}: They use fastText tool to generate word embeddings from Wikipedia\footnote{https://www.wikipedia.org/}. The format is the same at Kyubyong's, but their embedding is the vector of 300 dimension, and they have about 200k words

\subsection{Model}

Based on state-of-the-art methods for NER, BLSTM-CNN-CRF is the end-to-end deep neural network model that achieves the best result on F-score \cite{16}. Therefore, we decide to conduct the experiment on this model and analyze the errors.

We run experiment with the Ma and Hovy (2016) model \cite{8}, source code provided by (Motoki Sato)\footnote{The code of the BLSTM-CNN-CRF for NER systems are available at https://github.com/aonotas/deep-crf}  and analysis the errors from this result. Before we decide to analysis on this result, we have run some other methods, but this one with Vietnamese pre-trained word embeddings provided by Kyubyong Park obtains the best result. Other results are shown in the Table 2.
\section{Experiment and Results}
Table 2 shows our experiments on two models with and without different pre-trained word embedding – KP means the Kyubyong Park’s pre-trained word embeddings and EG means Edouard Grave’s pre-trained word embeddings.

\begin{table}
\label{tb2}
\caption{F1 score of two models with different pre-trained word embeddings}
\begin{center}
\begin{tabular}{ll}
\hline\rule{0pt}{12pt}
Model	&F1 (\%)\\[2pt]
\hline\rule{0pt}{12pt}Bi-LSTM-CRF (no word embedings) 	&84.87\\
Bi-LSTM-CRF (KP word embedings) 	&86.69\\
Bi-LSTM-CRF (EG word embedings) 	&85.80\\
Bi-LSTM-CNN-CRF (no word embedings)	&84.31\\
Bi-LSTM-CNN-CRF (KP word embedings)	&86.87\\
\end{tabular}
\end{center}
\end{table}

We compare the outputs of BLSTM-CNN-CRF model (predicted) to the annotated data (gold) and analyzed the errors.
Table 3 shows perfomance of the BLSTM-CNN-CRF model. In our experiments, we use three evaluation parameters (precision, recall, and F1 score) to access our experimental result. They will be described as follow in Table 3. The "correctNE", the number of correct label for entity that the model can found. The "goldNE", number of the real label annotated by annotator in the gold data. The "foundNE", number of the label the model find out (no matter if they are correct or not).

\begin{equation}Recall = 
  \frac{correctNE×100}{goldNE}
\end{equation}

\begin{equation}Precision = 
  \frac{correctNE×100}{foundNE}
\end{equation}

\begin{equation}F1 = 
  \frac{2×Precision×Recall}{Precision+Recall}
\end{equation}

\begin{table}
\label{tb3}
\caption{Performances of LSTM-CNN-CRF on the Vietnamese NER corpus}
\begin{center}
\begin{tabular}{llll}
\hline\rule{0pt}{12pt}Tag name	&Precision (\%)	&Recall (\%)	&F1(\%)
\\[2pt]
\hline\rule{0pt}{12pt}All Result	&87.70&	85.71&	86.70\\
LOC	&87.63&	86.87&	87.25\\
MISC&	97.44	&77.55&	86.36\\
PER&	90.15&	91.27&	90.71\\
ORG	&71.23	&55.11	&62.14\\

\end{tabular}
\end{center}
\end{table}
In Table 3 above, we can see that recall score on ORG label is lowest. The reason is almost all the ORG label on test file is name of some brands that do not appear on training data and pre-trained word embedding. On the other side, the characters inside these brand names also inside the other names of person in the training data. The context from both side of the sentence (future- and past-feature) also make the model "think" the name entity not as it should be.

Table 4 shows that the biggest number of errors is \textbf{No extraction}. The errors were counted by using logical sum (OR) of the gold labels and predicted labels (predicted by the model). The second most frequent error was \textbf{Wrong tag} means the model extract it's a NE but wrong tag.

\subsection{Error analysis on gold data}

First of all, we will compare the predicted NEs to the gold NEs (Fig. 1). Table 4 shows the summary of errors by types based on the gold labels, the "correct" is the number of gold tag that the model predicted correctly, "error" is the number of gold tag that the model predicted incorrectly, and "total" is sum of them. Four columns next show the number of type errors on each label.

Table 5 shows that Person, Location and Organization is the main reason why \textbf{No extraction} and \textbf{Wrong tag} are high.

After analyzing based on the gold NEs, we figure out the reason is:
\begin{itemize}
\item Almost all the NEs is wrong, they do not appear on training data and pre-trained embedding. These NEs vector will be initial randomly, therefore, these vectors are poor which means have no semantic aspect.\newline
\item The "weird" ORG NE in the sentence appear together with other words have context of PER, so this "weird" ORG NE is going to be label at PER.

For example:

\textbf{gold data}: \foreignlanguage{vietnam}{VĐV được xem là đầu\_tiên ký hợp\_đồng quảng\_cáo là võ\_sĩ ⟨PER⟩ Trần Quang Hạ ⟨PER⟩ sau khi đoạt HCV taekwondo Asiad ⟨LOC⟩ Hiroshima ⟨LOC⟩.}\newline
(The athlete is considered the first to sign a contract of boxing Tran Quang Ha after winning the gold medal Asiad Hiroshima)

\textbf{predicted data}: \foreignlanguage{vietnam}{…là võ\_sĩ ⟨PER⟩Trần Quang Hạ⟨PER⟩ sau khi đoạt HCV taekwondo Asiad ⟨PER⟩Hiroshima⟨PER⟩.}\newline

\item Some mistakes of the model are from training set, for example, anonymous person named "P." appears many times in the training set, so when model meets "P." in context of "P. 3 \foreignlanguage{vietnam}{Quận 9}" (Ward 3, District 9) – "P." stands for \foreignlanguage{vietnam}{"Phường"} (Ward) model will predict "P." as a PER.

Training data: \foreignlanguage{vietnam}{nếu ⟨PER⟩P.⟨PER⟩ có ở đây – (If P. were here)}
Predicted data: \foreignlanguage{vietnam}{⟨PER⟩P. 3⟨PER⟩, Gò\_vấp – (Ward 3, Go\_vap District)}
\end{itemize}

\begin{table}
\label{tb4}
\caption{Summary of error results on gold data}
\begin{center}
\begin{tabular}{lll}
\hline\rule{0pt}{12pt}Error type	&Number (NE)	&Rate (\%)
\\[2pt]
\hline\rule{0pt}{12pt}No extraction	&142	&33.18\\
Wrong tag 	&112	&26.17\\
Wrong range	&100	&23.36\\
Wrong range and tag	&74	&17.29\\
All errors& 	428	&100\\

\end{tabular}
\end{center}
\end{table}

\begin{table}
\label{tb5}
\caption{Summary of detailed error results on gold data}
\begin{center}
\begin{tabular}{llllcccc}
\hline\rule{0pt}{12pt}
Tags&Correct&Error&Total&No Extraction&Wrong Tag&Wrong Range&Wrong Range \& Tag
\\[2pt]
\hline\rule{0pt}{12pt}Person	&1181	&113	&1294	&51	&32	&24	&6\\
Location	&1198	&181	&1377	&54	&39	&59	&29\\
Org	&151	&123	&274	&31	&41	&17	&34\\
MISC	&38	&11	&49	&6&	0&	0&	5\\
All Tags&	2566&	428	&2994	&142	&112	&100	&74\\
\end{tabular}
\end{center}
\end{table}

\subsection{Analysis on predicted data}
Table 6 shows the summary of errors by types based on the predicted data. After analyzing the errors on predicted and gold data, we noticed that the difference of these errors are mainly in the \textbf{No anotation} and \textbf{No extraction}. Therefore, we only mention the main reasons for the \textbf{No anotation}:

Most of the wrong labels that model assigns are brand names (Ex: Charriol, Dream, Jupiter, ...), words are abbreviated \foreignlanguage{vietnam}{(XKLD – xuất khẩu lao động (labour export))}, movie names, … All of these words do not appear in training data and word embedding. Perhaps these reasons are the followings:\newline
\begin{itemize}
\item The vectors of these words are random so the semantic aspect is poor.\newline
\item The hidden states of these words also rely on past feature (forward pass) and future feature (backward pass) of the sentence. Therefore, they are assigned wrongly because of their context.\newline
\item These words are primarily capitalized or all capital letters, so they are assigned as a name entity. This error is caused by the CNN layer extract characters information of the word.
\end{itemize}

\begin{table}
\label{tb6}
\caption{Summary of error results on predicted data}
\begin{center}
\begin{tabular}{lll}
\hline\rule{0pt}{12pt}
Error type	&Number (NE)	&Rate (\%)
\\[2pt]
\hline\rule{0pt}{12pt}Wrong tag 	&113&	31.48\\
Wrong range	&88	&24.51\\
Wrong range and tag	&69	&19.22\\
No annotation	&89&	24.79\\
All errors& 	359	&100\\

\end{tabular}
\end{center}
\end{table}

\begin{table}
\label{tb7}
\caption{Summary of detailed error results on predicted data}
\begin{center}
\begin{tabular}{llllcccc}
\hline\rule{0pt}{12pt}
Tags&Correct&Error&Total&No Annotation&Wrong Tag&Wrong Range&Wrong Range \& Tag
\\[2pt]
\hline\rule{0pt}{12pt}Person	&1181&	129&	1310&	40&	52	&20&	17\\
Location	&1198	&169	&1367	&26	&54	&53	&36\\
Org	&151	&60&	212&	22&	7&	15&	16\\
MISC	&38	&1&	39&	1&	0&	0&	0\\
All Tags&	2566	&359&	2928&	89&	113&	88&	69\\
\end{tabular}
\end{center}
\end{table}

Table 7 shows the detail of errors on predicted data where we will see number kind of errors on each label.

\subsection{Errors of annotators}
After considering the training and test data, we realized that this data has many problems need to be fixed in the next run experiments. The annotators are not consistent between the training data and the test data, more details are shown as follow:\newline

\begin{itemize}
\item The organizations are labeled in the train data but not labeled in the test data:\newline
\textbf{Training data}: \foreignlanguage{vietnam}{⟨ORG⟩ Sở Y\_tế ⟨ORG⟩} (Department of Health)\newline \textbf{Test data}: {\foreignlanguage{vietnam}{Sở Y\_tế} (Department of Health)}\newline
\textbf{Explanation}: \foreignlanguage{vietnam}{"Sở Y\_tế"} in train and test are the same name of organization entity. However the one in test data is not labeled.\newline

\item The entity has the same meaning but is assigned differently between the train data and the test:
\newline\textbf{Training data}: \foreignlanguage{vietnam}{⟨MISC⟩ người Việt ⟨MISC⟩} (Vietnamese people)
\newline \textbf{Test data}: \foreignlanguage{vietnam}{dân ⟨LOC⟩ Việt ⟨LOC⟩} (Vietnamese people)\newline
\textbf{Explanation}: \foreignlanguage{vietnam}{Both "người Việt" in train data and "dân Việt" in test data} are the same meaning, but they are assigned differently.\newline

\item The range of entities are differently between the train data and the test data:
\newline \textbf{Training data}: \foreignlanguage{vietnam}{⟨LOC⟩ làng Atâu ⟨LOC⟩ (Atâu village)}
\newline \textbf{Test data}: \foreignlanguage{vietnam}{làng ⟨LOC⟩ Hàn\_Quốc ⟨LOC⟩ (Korea village)}
\newline 
\textbf{Explanation}: The two villages differ only in name, but they are labeled differently in range
\newline 

\item Capitalization rules are not unified with a token is considered an entity:
\newline \textbf{Training data}: \foreignlanguage{vietnam}{⟨ORG⟩ Công\_ty Inmasco ⟨ORG⟩ (Inmasco Company)}
\newline \textbf{Training data}: \foreignlanguage{vietnam}{công\_ty con (Subsidiaries)}
\newline \textbf{Test data}: \foreignlanguage{vietnam}{công\_ty ⟨ORG⟩ Yeon Young Entertainment ⟨ORG⟩} (Yeon Young Entertainment company)\newline
\textbf{Explanation}: If it comes to a company with a specific name, it should be labeled \foreignlanguage{vietnam}{⟨ORG⟩ Công\_ty Yeon Young Entertainment ⟨ORG⟩ with "C" in capital letters.}
\end{itemize}

\section{Conclusion}
In this paper, we have presented a thorough study of distinctive error distributions produced by Bi-LSTM-CNN-CRF for the Vietnamese language. This would be helpful for researchers to create better NER models.

Based on the analysis results, we suggest some possible directions for improvement of model and for the improvement of data-driven NER for the Vietnamese language in future:
\begin{enumerate}
\item The word at the begin of the sentence is capitalized, so, if the name of person is at this position, model will ignore them (no extraction). To improve this issue, we can use the POS feature together with BIO format (Inside, Outside, Beginning) \cite{4} at the top layer (CRF).
\item If we can unify the labeling of the annotators between the train, dev and test sets. We will improve data quality and classifier.
\item It is better if there is a pre-trained word embeddings that overlays the data, and segmentation algorithm need to be more accurately.
\end{enumerate}

%

\end{document}